\documentclass[letterpaper, 10 pt, conference]{ieeeconf}  

\IEEEoverridecommandlockouts                              

\usepackage{amssymb,amsfonts,amsmath,amsthm,amscd}
\usepackage{epsfig,graphics,psfrag}

\newtheorem{lemma}{Lemma}

\newtheorem{conj}{Conjecture}

\def\prob{\mathbb P}

\def\E{\mathbb E}
\def\de{\rm d}

\def\|{\big|\big|}

\def\Tree{{\sf T}}

\def\ux{\underline{x}}
\def\u0t{{\tt \underline{0}}}
\def\0t{{\tt 0}}
\def\1t{{\tt 1}}

\def\ind{{\mathbb I}}

\def\R{{\mathbb R}}
\def\da{{\partial a}}
\def\di{{\partial i}}
\def\dpi{{\partial_+ i}}
\def\dmi{{\partial_- i}}
\def\dmj{{\partial_- j}}
\def\dpmi{{\partial_\pm i}}
\def\Ball{{\sf B}}
\def\poisson{{\sf Poisson}}
\def\ed{\stackrel{{\rm d}}{=}}

\def\hw{{\mathfrak h}}
\def\uw{{\mathfrak u}}
\def\imp{{\tt I}}
\def\nimp{{\tt 0}}
\def\sBPd{{\mbox{\footnotesize BPd}}}
\def\stree{{\mbox{\footnotesize\rm tree}}}
\def\hd{\widehat{h}}
\def\ud{\widehat{u}}
\def\hphi{\widehat{\phi}}
\def\halpha{\widehat{\alpha}}
\def\sp{{\mbox{\footnotesize spin}}}

\newcommand{\comment}[1]{}

\title{\Large \bf Solving Constraint Satisfaction Problems through\\
Belief Propagation-guided decimation}

\author{Andrea Montanari, Federico Ricci-Tersenghi and Guilhem Semerjian
\thanks{A.~Montanari is with Departments of Electrical Engineering 
and Statistics, Stanford University, 
{\tt\small montanari@stanford.edu},
F.~Ricci-Tersenghi is with Dipartimento di Fisica, Universit\`a di Roma 
La Sapienza, G.~Semerjian is with Laboratoire de Physique
Th\'eorique de l'Ecole Normale, Paris}}

\begin{document}

\maketitle
\thispagestyle{empty}
\pagestyle{empty}
\begin{abstract}
Message passing algorithms have proved surprisingly successful in solving
hard constraint satisfaction problems on sparse random graphs.
In such applications, variables are fixed sequentially to 
satisfy the constraints. Message passing is run after each step.
Its outcome provides an heuristic to make choices at next step. 
This approach has been referred to as `decimation,' with reference 
to analogous procedures in statistical physics.

The behavior of decimation procedures is poorly understood. 
Here we consider a simple randomized decimation algorithm 
based on belief propagation (BP), and analyze its behavior on 
random $k$-satisfiability formulae. In particular, 
we propose a tree model for its analysis and we conjecture 
that it provides asymptotically exact predictions in the limit of 
large instances. This conjecture is confirmed by numerical simulations.
\end{abstract}
%
%
\section{Introduction}

An instance of a constraint satisfaction problem \cite{Creignou} consists of
$n$ variables $\ux = (x_1,\dots,x_n)$ and $m$ constraints among them. 
Solving such an instance amounts to finding an assignment of the variables 
that satisfies all the constraints, or proving that no such assignment 
exists. A remarkable example in this class is provided by $k$-satisfiability,
where variables are binary, $x_i\in \{\0t,\1t\}$, and each 
constraint requires $k$ of the variables to be different 
from a specific $k$-uple. Explicitly, the $a$-th constraint (clause),
$a\in[m]\equiv\{1,\dots,m\}$ is specified by $k$ variable indexes
$i_1(a), \dots,i_k(a)\in[n]$, and $k$ bits $z_1(a),\dots, z_k(a)
\in\{\0t,\1t\}$. Clause $a$ is satisfied by assignment $\ux$ if and only if 
$(x_{i_1(a)},\dots,x_{i_k(a)})\neq (z_1(a),\dots,z_k(a))$.

\begin{figure}[t]
\center{\includegraphics[width=4cm]{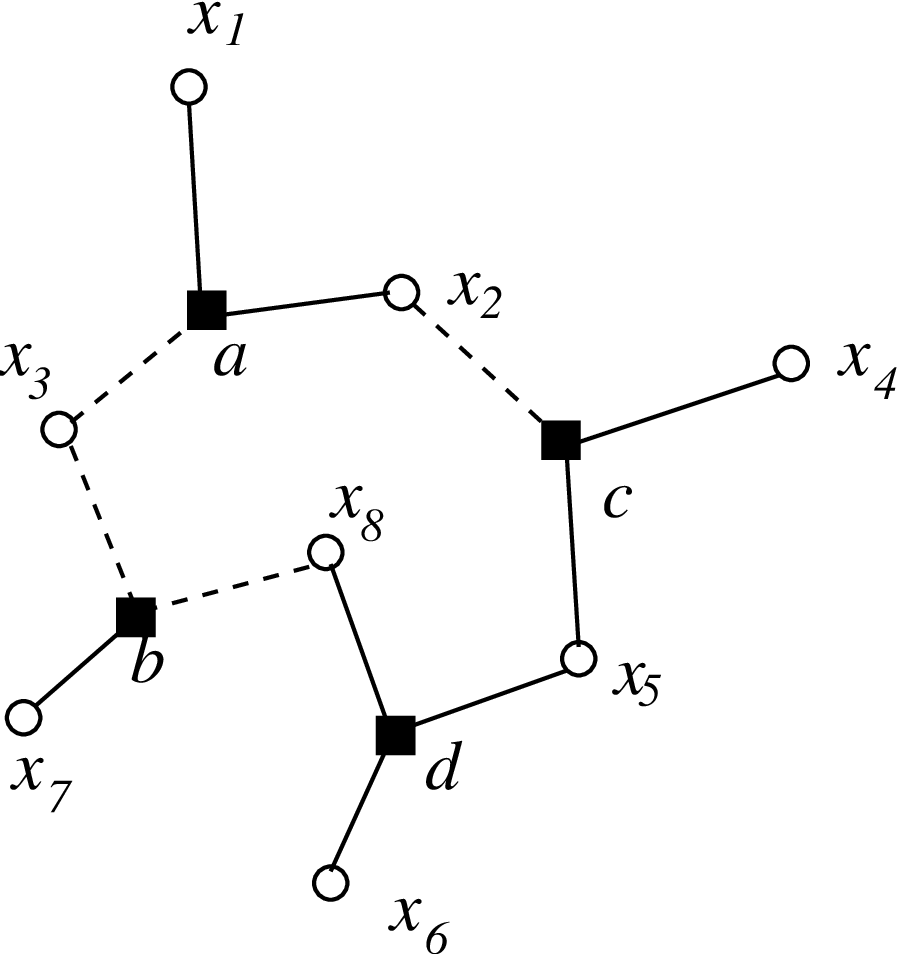}}
\caption{Factor graph of a small 3-SAT instance. 
Continuous edges correspond to $z_j(a) =\0t$, and dashed ones
to $z_j(a) =\1t$. The corresponding Boolean formula reads
$(x_1\vee x_2\vee\overline{x}_3)\wedge (\overline{x}_2\vee x_4\vee x_5)\wedge
(x_5\vee x_6\vee x_8)\wedge (\overline{x}_3\vee x_7\vee\overline{x}_8)$.}
\label{fig:Factor}
\vspace{-4mm}
\end{figure}

A constraint satisfaction problem admits a natural factor graph~\cite{Factor} 
representation, cf. Fig.~\ref{fig:Factor}. 
Given an instance, each variable can be associated to a 
variable node, and each constraint to a factor node. Edges
connect factor node $a\in F\equiv [m]$ to those variable nodes 
$i\in V\equiv [n]$ such that the $a$-th constraint depends in a non-trivial
way on variable $x_i$.
For instance, in the case of $k$-satisfiability, clause $a$
is connected to variables $i_1(a),\dots,i_k(a)$. If the resulting 
graph is sparse, fast message passing algorithms can be defined on it.

Although constraint satisfaction problems are generally NP-hard,
a large effort has been devoted to the development of efficient heuristics.
Recently,  considerable progress has been achieved in 
building efficient `incomplete solvers' \cite{Selman}. 
These are algorithms that look for a 
solution but,  if they do not find one, cannot prove that the problem is
unsolvable. A particularly interesting class is provided by 
\emph{message passing-guided decimation} procedures. These consist
in iterating the following steps:
\begin{enumerate}
\item Run a message passing algorithm. 
\item Use the result to choose a variable index $i\in V$,
and a value $x_i^*$ for the corresponding variable.
\item Replace the constraint satisfaction problem with the one
obtained by fixing $x_i$ to $x_i^*$.
\end{enumerate}
The iteration may stop for two reasons. In the first case a contradiction is
produced:  the same variable $x_i$ appears in two constraints whose 
other arguments have already been fixed, and that are satisfied by distinct 
values of $x_i$. If this does not happen, the iteration stops
only when all the variables are fixed and a solution is found.
Notice that earlier algorithms, such as unit clause propagation (UCP)
\cite{Franco,Myopic} did not used message passing in step 2, and were not 
nearly as effective.

Random constraint satisfaction problems are a useful testing ground
for new heuristics. For instance, \emph{random $k$-satisfiability}
is the distribution over $k$-SAT formulae defined by picking a
formula uniformly at random among all the ones including 
$m$ clauses over $n$ variables. Decimation procedures of the type sketched 
above proved particularly successful in this context. In particular 
\emph{survey propagation}-guided decimation~\cite{Mezard,Zecchina} outperformed the 
best previous heuristics based on stochastic local search \cite{Selman}. 
More recently \emph{belief propagation}-guided decimation
was shown empirically to have good performances as well
\cite{OurPNAS}.

Unfortunately, so far there exists no analysis of  message-passing guided
decimation. Our understanding almost entirely relies on simulations, 
even for random instances. Consequently the comparison among different 
heuristics, as well as the underpinnings of their effectiveness are
somewhat unclear. In this paper we define a simple class 
of randomized message passing-guided decimation algorithms, 
and present a technique for analyzing them on random instances. 
The technique is based on the identification of a process on infinite trees 
that describes the evolution of the decimation algorithm. 
The tree process is then analyzed 
through an appropriate generalization of density evolution \cite{MCT}.
Our approach is close in
spirit to the one of~\cite{Luby}.
While it applies to a large class of random constraint satisfaction
problems (including, e.g. coloring of random graphs),
for the sake concreteness, we will focus  on random $k$-SAT.

We expect the tree process to describe exactly the algorithm behavior in 
the limit of large instances, $n\to\infty$. While we could not 
prove this point, numerical simulations convincingly support this conjecture.
Further, non-rigorous predictions based on tree calculations have been 
repeatedly successful in the analysis of random $k$-satisfiability.
This approach goes under the name of `cavity method' in statistical 
mechanics~\cite{Mezard}. 

The paper is organized as follows. Section \ref{sec:Back}  
contains some necessary background and notation on random 
$k$-SAT as well as a synthetic discussion of related work.
In Section \ref{sec:Decimation}
we define the decimation procedure that we are going to 
analyze. We further  provide the basic intuition behind the definition of
the tree model.
The latter is analyzed in Section \ref{sec:Tree}, and the predictions 
thus derived are compared with numerical simulations in Section
\ref{sec:Numerical}. Finally, some conclusions and suggestions for future work
are presented in Section \ref{sec:Conclusions}.
Proofs of several auxiliary lemmas are omitted from this extended 
abstract and deferred to technical appendices.
%
%
\section{Random $k$-SAT and message passing: Background and related work}
\label{sec:Back}

As mentioned above, \emph{random $k$-SAT} refers to the uniform distribution 
over $k$-SAT instances with $m$ constraints over $n$ variables. More 
explicitly, each constraint is drawn uniformly at random among the 
$2^k\binom{n}{k}$ possible ones. We are interested here in the limit 
$n,m\to\infty$ with $m/n=\alpha$ fixed. 

Consider the factor graph $G$ of a random $k$-SAT formula,
endowed with the graph-theoretic distance. Namely, the 
distance of two variable nodes $d(i,j)$ is the length of the shortest path 
leading from $i$ to $j$ on $G$.
It is well known \cite{Luczak} 
that, in the large size limit, any finite neighborhood
of a random node $i$ converges in distribution to a well defined random tree.
This observation will be the basis of our tree analysis of the decimation
process, and is therefore worth spelling it out in detail.
Let $\Ball(i,\ell)$ be the subgraph induced by all the vertices
$j\in G$, such that $d(i,j)\le \ell$. 
Then $\Ball(i,\ell)\stackrel{{\rm d}}{\rightarrow} \Tree(\ell)$ as 
$n\to \infty$, where  $\Tree(\ell)$ is the random rooted
(factor) tree defined recursively as follows. 
For $\ell=0$, $\Tree(\ell)$  is the graph containing
a unique variable node. For any $\ell \ge 1$, start by a single variable node
(the root) and add $l\ed \poisson(\alpha k)$ clauses, each one
including the root and $k-1$ new variables (first generation variables).
If $\ell \ge 2$,
generate an independent copy of $\Tree(\ell-1)$ for each variable node
in the first generation and attach it to them. The values $z_j(a)$
that violate clause $a$ are independently chosen in $\{\0t,\1t\}$
with equal probability.
It is easy to see that the limit object $\Tree(\infty)$ is well defined and 
is an infinite tree with positive probability if $\alpha> 1/k(k-1)$.

We let 
$\alpha_{\rm s}(k)$ be the largest value of $\alpha$ such that random
$k$-SAT instances admit with high probability a solution.
It is known~\cite{AP04} that $\alpha_{\rm s}(k) = 2^{k}\log 2-O(k)$.
A sharp conjecture on the value of $\alpha_{\rm s}(k)$ has been put
forward in~\cite{Mezard} on the basis of statistical physics 
calculations,
implying $\alpha_{\rm s}(k) \approx 4.267$, $9.93$, $21.12$ for (respectively)
$k=3,4,5$ and $\alpha_{\rm s}(k) = 2^{k}\log 2-\frac{1}{2}(1+\log 2)+
O(2^{-k})$ for large $k$~\cite{Mertens}.

Simple heuristics have been analyzed thoroughly \cite{Myopic} and proved to 
find a solution with probability bounded away from $0$
if $\alpha\le {\rm const}\, 2^{k}/k$. Here the 
proportionality constant depends on the specific heuristic.

To the best of our knowledge, the first application of message passing 
algorithms to $k$-satisfiability is reported in \cite{Pumphrey}. In this
early study BP was mostly applied in a \emph{one-shot fashion} (as in
iterative decoding of sparse graph codes \cite{MCT}), without decimation.
By this we mean that belief propagation is run, and resulting marginal
probabilities are used to guess the values of all variables at once.
However the probability of success of the one-shot algorithm is 
exponentially small: there are $\Theta(n)$ isolated constraints,
whose variables have non-trivial marginal probabilities, each of them is
hence violated with finite probability in the one-shot assignment.

Statistical mechanics methods allowed to derive a very precise picture 
of the solution set~\cite{BiMoWe,Mezard,OurPNAS}. 
This inspired a new message passing algorithm dubbed 
\emph{survey propagation}~\cite{Zecchina}. In conjunction with decimation, 
this algorithm allowed to solve random instances of unprecedentedly 
large sizes, 
in difficult regimes of $\alpha$ and $k$. 

A natural way of introducing belief propagation for $k$-satisfiability 
is to consider the uniform distribution over solutions (assuming their 
existence). Let us
denote by $\da =\{i_1(a),\dots,i_k(a)\}$ the set of variable nodes on
which the $a$-th constraint effectively depends, for any subset $U$
of the variable nodes their partial assignment $\ux_U=\{x_i \, | \, i \in U\}$,
and $w_a(\ux_\da) = \ind\big\{ (x_{i_1(a)},\dots,x_{i_k(a)})\neq (z_{1}(a),
\dots,z_k(a))\big\}$ the indicator function of the event 'clause $a$ is 
satisfied.' The uniform distribution over the solutions can thus be written
\begin{equation}
\mu(\ux) = \frac{1}{Z}\, \prod_{a\in F}\, w_a(\ux_\da) \ .
\label{eq:Uniform}
\end{equation}
In~\cite{MontanariShah} it was proved that for
$\alpha\le (2\log k)/k\, [1+o(1)]$, BP computes good approximations
of the marginals of $\mu$, irrespective of its initialization. It is clear 
from empirical studies~\cite{MPEmpirical,Maneva}
that the `worst case'  argument used in this estimate (and in other
papers on belief propagation \cite{TJ99,GamarnikColoring}) 
is far too pessimistic. 

In Ref.~\cite{MosselPlanted} a simple message passing algorithm,
warning propagation (see below), was analyzed for a modified (`planted')
ensemble of random formulae. The algorithm was proved to converge and find 
solutions for large enough density $\alpha$ 
(see also~\cite{unif_sat,unif_sat2}). 
Both the ensemble and the algorithm are quite different from the ones 
treated in this paper.

Further, the definition and analysis of a `Maxwell decoder'
in~\cite{Maxwell1,Maxwell2}, is closely related to the approach in this
paper. Let us recall that the Maxwell decoder was a (mostly conceptual)
algorithm for implementing maximum likelihood decoding of 
LDPC codes over the erasure channel. The treatment in~\cite{Maxwell1,Maxwell2}
applies almost verbatim to a simple constraint satisfaction 
problem known as XORSAT. The generalization in the present paper 
is analogous to the one from the erasure to a general binary memoryless
symmetric channel.

Finally, let us mention that BP decimation can be an interesting
option in engineering applications, as demonstrated empirically in the case 
of lossy source coding \cite{ManevaWainwright,Ciliberti}.
%
%
\section{A simple decimation procedure}
\label{sec:Decimation}

\subsection{Belief propagation}

Let us recall the definition of BP for our specific setup 
(\cite{Factor,Pearl} are general references). BP is a
message passing algorithm: at each iteration messages are
sent from variable nodes to neighboring clause nodes and vice versa. 
To describe the message update equations, we need some more notation.
As in the case of factor nodes, we shall call $\di$ the set of factors that
depends on the variable $x_i$. 
If $i\in \da$, say $i=i_l(a)$, we denote $z(i,a)=z_l(a)$
the value of $x_i$ which does not satisfy the $a$-th clause.
For a pair of adjacent variable ($i$) and factor ($a$) nodes 
(i.e. $i \in \da$), let us call $\dpi(a)$ (resp. $\dmi(a)$) the set of
factor nodes adjacent to $i$, distinct from $a$, that agrees (resp. disagrees)
with $a$ on the satisfying value of $x_i$. 
In formulae, $\dpi(a) = \{b \in \di \setminus a | z(i,b) =z(i,a) \}$ and
$\dmi(a) = \{b \in \di | z(i,b) = 1- z(i,a) \}$.

It is convenient to use log-likelihood notations for messages as is
done in iterative decoding \cite{MCT}, with two caveats:
$(1)$ We introduce a factor $1/2$ to be consistent with physics notation;
$(2)$ The message from variable node $i$ to factor node $a$
corresponds to the log-likelihood for $x_i$ to satisfy/not-satisfy  clause $a$
(rather than to be $\0t/\1t$).

Let 
$\{h^{(r)}_{i\to a}\}$, $\{u^{(r)}_{a\to i}\}$ denote the messages
that are passed at time $r$ along the directed edges
$i\to a$ and $a\to i$, for $i\in V$, and $a\in F$.
The update equations read
\begin{eqnarray}
h_{i\to a}^{(r+1)} & = & \sum_{b\in\dpi(a)}u^{(r)}_{b\to i}
-\sum_{b\in\dmi(a)}u^{(r)}_{b\to i}\,
, \label{eq:BPUpdate1}\\
u^{(r)}_{a\to i} & = & f(\{h^{(r)}_{j\to a};j\in\da\backslash i\})\, ,
\label{eq:BPUpdate2}
\end{eqnarray}
where we define the function $f:\R^{k-1}\to\R$ as
\begin{equation}
f(h_1,\dots,h_{k-1}) = -\frac{1}{2}\log\left\{1-
\frac{1-\epsilon}{2^{k-1}}\prod_{i=1}^{k-1}(1-\tanh h_i)\right\}\, ,
\label{eq:FDefinition}
\end{equation}
with $\epsilon=0$ (this parameter is introduced for the discussion in
Sec.~\ref{sec:Numerical}). 

For $i\in V$, let $\dpi$ be the subset of clauses that are satisfied
by $x_i = \0t$, and $\dmi$ the subset satisfied by $x_i=\1t$. Then
the BP estimate for the marginal of $x_i$ under the measure $\mu(\,\cdot\,)$ 
is $\nu^{(r)}_i(x_i)$, where
\begin{eqnarray}
\nu^{(r)}_i(\0t/\1t) &=& \frac{1 \pm \tanh h_i^{(r)}}{2} \ , \nonumber \\
h_i^{(r)} &=& \sum_{a\in\dpi}u^{(r)}_{a\to i}
-\sum_{a\in\dmi}u^{(r)}_{a\to i}\, .
\label{eq:BPMarginal}
\end{eqnarray}
%
%
%
\subsection{Unit clause and warning propagation}

During the decimation procedure a subset $U$ of the variables 
are \emph{fixed} to specific values, collectively denoted as $\ux_U^*$.
This has some \emph{direct implications}. By this we mean
that for some other variables $x_j$, $j\in V\setminus U$, it follows
from `unit clause propagation' (UCP) that they take the same value in all
of the solutions compatible with the partial assignment $\ux_U^*$.
We will say that these variables are directly implied by the condition
$\ux_U =\ux_U^*$. Let us recall that unit clause propagation
corresponds to the following deduction procedure. 
For each of the fixed variables $x_i$, and each of the clauses $a$ 
it belongs to, the value $x_i^*$ can either satisfy clause $a$, or not.
In the first case clause $a$ can be eliminated from the factor graph. 
In the second a smaller clause with one less variable is implied. 
In both cases variable $x_i$ is removed. It can happen that the 
size of a clause gets reduced to $1$, through this procedure. In this case
the only variable belonging to the clause must take a definite value in order to
satisfy it. We say that such a variable is directly implied
by the fixed ones. Whenever a variable is directly implied,
its value can be substituted in all the clauses it belongs to, 
thus allowing further reductions.

The process stops for one of two reasons: $(1)$ All the fixed or directly 
implied variables have been pruned and no unit clause is present in the 
reduced formula. In this case we refer to all variables that appeared at some
point in a unit clause as \emph{directly implied} variables. 
$(2)$ Two unit clauses imply different values for the same variable.  
We will say that a \emph{contradiction} was revealed in this case:
no solution $\ux$ of the formula can verify the condition 
$\ux_U =\ux_U^*$.

A key element in our analysis is the remark that UCP admits a 
message passing description. The corresponding algorithm 
is usually referred to as \emph{warning propagation}
(WP)~\cite{Braunstein}. 
The WP messages (to be denoted as $\hw_{i\to a}^{(r)}$, 
$\uw_{a\to i}^{(r)}$) take values in $\{\imp,\nimp\}$.
The meaning of $\uw_{a\to i}^{(r)}=\imp$ (respectively 
$\uw_{a\to i}^{(r)}=\nimp$) is: `variable $x_i$ is (resp. is not)
directly implied
by clause $a$ to satisfy it.' For variable-to-factor messages,
the meaning of $\hw_{i\to a}^{(r)}=\imp$  (respectively
$\hw_{i\to a}^{(r)}=\nimp$) is: `variable $x_i$ is (resp. is not)
directly implied, through one of the clauses $b\in\di\setminus a$,  not 
to satisfy clause $a$.'

We want to apply WP to the case
in which a part of the variables have been fixed, namely $x_i=x^*_i$
for any $i\in U\subseteq V$.
In this case the WP rules read
\begin{eqnarray}
\hw_{i\to a}^{(r+1)} & = & \left\{\begin{array}{ll}
\imp & \begin{array}{l}
\mbox{if $\exists b\in \dmi(a)$ s.t. $\uw^{(r)}_{b\to i}=\imp$}\\
\mbox{or $i\in U$ and $x_i^*= z(i,a)$,}\end{array}\\
\nimp & \mbox{ otherwise,}\end{array}\right.\label{eq:WPUpdate1}\\
\uw^{(r)}_{a\to i} & = & \left\{\begin{array}{ll}
\imp & \mbox{ if  $\hw^{(r)}_{j\to a}=\imp$  $\forall j\in \da\setminus i$,}\\
\nimp & \mbox{ otherwise.}\end{array}\right.\label{eq:WPUpdate2}
\end{eqnarray}
In the following we shall always assume that WP is initialized with 
$\hw_{i\to a}^{(0)} = \nimp$, $\uw_{i\to a}^{(0)} = \nimp$
for each edge $(ia)\in E$. 
It is then easy to prove that messages
are monotone in the iteration number (according to the ordering $\nimp<\imp$).
In particular the WP iteration converges in at most $O(n)$ iterations.
We denote $\{\uw_{a\to i}^{(\infty)}\}$ the corresponding fixed point messages,
and say that $i\in V\setminus U$ is \emph{WP-implied} by the fixed variables
if there exist $a\in\di$ such that $\uw_{a\to i}^{(\infty)}=\imp$. 
Then the equivalence between UCP and WP can be stated in the form below.
\begin{lemma}
Assume a partial assignment $\ux_U^*$
to be given for $U\subseteq V$. Then 
\begin{enumerate}
\item The fixed point WP messages
$\{\uw_{a\to i}^{(\infty)}\}$ do not depend on the order of the
WP updates (as long as any variable is updated an \emph{a priori}
unlimited number of times).
\item  $i\in V\setminus U$
is directly implied iff it is WP-implied.
\item UCP encounters a contradiction iff 
there exists $i\in V$, $a\in\dpi$, $b\in\dmi$ 
such that $\uw_{a\to i}^{(\infty)}=\uw_{b\to i}^{(\infty)}=\imp$.
\end{enumerate}
\end{lemma}

For the clarity of what follows let us emphasize the terminology
of \emph{fixed} variables (those in $U$) and of 
\emph{directly implied} variables 
(not in $U$, but implied by $\ux_U^*$ though UCP or WP).
Finally, we will call \emph{frozen} variables
the union of fixed and directly implied ones, and denote the set of
frozen variables by $W\subseteq V$. 
%
%
\subsection{Decimation}

The BP-guided decimation algorithm is defined by the
pseudocode of Table \ref{tab:Algo}.
\begin{table}
{\normalsize
\begin{tabular}{ll}
\hline
\multicolumn{2}{l}{BP-Decimation ($k$-SAT instance $G$)}\\
\hline
1: & initialize BP messages $\{h_{i\to a}=0, u_{a\to i}=0\}$; \\
2: & initialize WP messages $\{\hw_{i\to a}=\nimp, 
\uw_{a\to i}=\nimp\}$;\\
3: & initialize $U=\emptyset$;\\
4: & {\bf for} $t=1,\dots,n$\\
5: & \hspace{0.5cm}run BP until the stopping criterion is met;\\ 
6: & \hspace{0.5cm}choose $i\in V\setminus U$ uniformly at random;\\
7: & \hspace{0.5cm}compute the  BP marginal $\nu_i(x_i)$;\\
8: & \hspace{0.5cm}choose $x_i^*$ distributed according to $\nu_i$;\\
9: & \hspace{0.5cm}fix $x_i = x_i^*$ and set $U\leftarrow U\cup \{i\}$;\\
10: & \hspace{0.5cm}run WP until convergence;\\
11:& \hspace{0.5cm}if a contradiction is found, {\bf return} FAIL;\\
12:& {\bf end}\\
13:& {\bf return} $\ux^*$.\\
\hline
\end{tabular}}
\caption{The Belief Propagation-guided decimation algorithm.}\label{tab:Algo}
\vspace{-5mm}
\end{table}
There are still a couple of elements we need to specify. First of
all, how the BP equations (\ref{eq:BPUpdate1}), (\ref{eq:BPUpdate2}) are 
modified when a non-empty subset $U$ of the variables is fixed.
One option would be to eliminate these variables from the factor graph,
and reduce the clauses they belong to accordingly. A 
simpler approach consists in modifying Eq.~(\ref{eq:BPUpdate1})
when $i\in U$. Explicitly, if the chosen value $x_i^*$ satisfies 
clause $a$, then we set $h_{i\to a}^{(r+1)} = +\infty$. If
it does not, we set  $h_{i\to a}^{(r+1)} = -\infty$.

Next, let us stress that, while WP is run until convergence, a
not-yet defined `stopping criterion' is used for BP. This will be 
precised in Section \ref{sec:Numerical}. Here we just say that 
it includes a maximum iteration number $r_{\rm max}$, which is kept of
smaller order than $O(n)$.

The algorithm complexity is therefore naively $O(n^3r_{\rm max})$.
It requires $n$ cycles, each involving: $(1)$ at most 
$r_{\rm max}$ BP iterations of $O(n)$ complexity and $(2)$ at most $n$
WP iterations of complexity $O(n)$.
It is easy to reduce the complexity to $O(n^2r_{\rm max})$ by updating
WP in sequential (instead of parallel) order, as in  UCP.
Finally, natural choice (corresponding to the assumption that BP converges 
exponentially fast) is to take $r_{\rm max}= O(\log n)$,
leading to  $O(n^2\log n)$ complexity.

In practice  WP converges after a small 
number of iterations, and the BP updates are the most expensive part
of the algorithm. This could be reduced further by using the fact that 
fixing a single variable should produce only a small change in the messages.
Ref.~\cite{Zecchina} uses this argument for a similar algorithm to argue that
$O(n\log n)$ time is enough. 
%
%
\subsection{Intuitive picture}
\label{sec:Intuitive}

Analyzing the dynamics of  BP-decimation seems extremely challenging.
The problem is that the procedure is not `myopic' \cite{Myopic},
in the sense that the
value chosen for variable $x_i$ depends on a large neighborhood of
node $i$ in the factor graph.
By analogy with myopic decimation algorithms 
one expects the existence of a critical value of the clause 
density $\alpha_{\sBPd}(k)$ such that the algorithm finds a solution 
with probability bounded away from $0$ for $\alpha<\alpha_{\sBPd}(k)$,
while it is unsuccessful with high probability for 
$\alpha>\alpha_{\sBPd}(k)$. Notice that, if the algorithm finds
a solution with positive probability, restarting it 
a finite number of times should\footnote{A caveat: here we are blurring 
the distinction between probability with respect to the formula 
realization and the algorithm realization.} 
yield a solution with probability arbitrarily close to 
$1$.

We shall argue in favor of this scenario and present an 
approach to analyze the algorithm evolution 
for $\alpha$ smaller than a \emph{spinodal point}
$\alpha_{\sp}(k)$. 
More precisely, our analysis allows to compute the asymptotic fraction 
of `directly implied' variables after any number of iterations.
Further, the outcome of this computation provides a strong indication 
that $\alpha_{\sp}(k)\le\alpha_{\sBPd}(k)$.  
Both the analysis, and the conclusion that 
$\alpha_{\sp}(k)\le\alpha_{\sBPd}(k)$ are confirmed by large scale numerical 
simulations. 


Our argument goes in two steps. 
In this section we show how to reduce
the description of the algorithm to a sequence of `static' problems. The
resolution of the latter will be treated in the next section.
Both parts rely on some assumptions on the asymptotic
behavior of large random $k$-SAT instances, that originate in
the statistical mechanics treatment of this problem \cite{Mezard,OurPNAS}.
We will spell out such assumptions along the way.

As a preliminary remark, notice that the two message passing algorithms 
play different roles in the BP-decimation procedure of Table
\ref{tab:Algo}. BP is used to estimate marginals of the uniform measure 
$\mu(\,\cdot\,)$ over solutions, cf.~Eq.~(\ref{eq:Uniform}), 
in the first repetition of the loop. 
In subsequent repetitions, it is used to compute marginals of 
the conditional distribution, given the current assignment 
$\ux_U = \ux^*_U$. These marginals are in turn used to 
choose the values $\{x_i^*\}$ of variables to be fixed.
WP is on the other hand used to check a necessary condition for the current
partial assignment to be consistent. Namely it checks if it induces a 
contradiction on directly implied variables. 
In fact, it could be replaced by UCP, and, in any case, it does not influence
the evolution of the partial assignment $\ux^*_U$.

Let us introduce some notation: $(i(1),i(2),\dots,i(n))$ is the order in which 
variables are chosen at step $5$ in the algorithm, $U_t=\{i(1),\dots,i(t) \}$
the set of fixed variables at the beginning of the $t+1$-th repetition of the loop,
and $W_t$ the frozen variables at that time (i.e. the union of $U_t$ and
the variables directly implied by $\ux^*_{U_t}$).

We begin the argument by considering an `idealized' version of the
algorithm where BP is replaced by a black box, that  
is able to return the exact
marginal distribution of the measure conditioned on the previous choices,
namely $\nu_i(\,\cdot\,) = \mu_{i|U}(\,\cdot\,|\ux_U^*)$. 
Let us point out two simple properties of this idealized algorithm.
First, it always finds a solution if the input formula is
satisfiable (this will be the case with high probability if we assume
$\alpha<\alpha_{\rm s}(k)$). In fact,
assume by contradiction that the algorithm fails.
Then,  there has been a last time $t$,
such that the $k$-SAT instance has at least one solution consistent
with the condition $\ux_{U_{t-1}} = \ux^*_{U_{t-1}}$, but no solution under the
additional constraint $x_{i}=x_{i}^*$ for $i=i(t)$. 
This cannot happen
because it would imply $\mu_{i|U_{t-1}}(x^*_{i}|\ux^*_{U_{t-1}}) = 0$, and 
if this is the case, we would not 
have chosen $x^*_i$ in step $8$ of the algorithm.

The second consequence is that the algorithm output configuration $\ux^*$ 
is a uniformly random solution. This follows from our assumption since
\begin{eqnarray*}
\prob\{\ux^*|i(\,\cdot\,)\} &=& \prod_{t=1}^n\nu_{i(t)}(x_{i(t)}^*) =\\
&=&\prod_{t=1}^n \mu_{i(t)|U(t-1)}(x^*_t|\ux^*_{U(t-1)}) = \mu(\ux^*)\, .
\end{eqnarray*}
Therefore, the distribution of the state of the idealized algorithm after 
any number $t$ of decimation steps can be described as follows. 
Pick a uniformly random
solution $\ux^*$, and a uniformly random subset of the variable indexes 
$U_t\subseteq V$, with $|U_t|=t$. Then fix the variables $i\in U_t$ 
to take value $x_i=x_i^*$, and discard the rest of the reference 
configuration $\ux^*$ (i.e. the bits $x_j^*$ for $j \notin U_t$).

We now put aside the idealized algorithm and consider  the effect 
of fixing the $t$-th variable $i(t)$ to $x_{i(t)}^*$.
Three cases can in principle arise: $(i)$ $x_{i(t)}$ was directly implied to 
be equal to $1-x_{i(t)}^*$ by $\ux^*_{U_{t-1}}$ and a contradiction 
is generated. We assume that BP is able to detect this direct 
implication and avoid such a trivial contradiction; 
$(ii)$ $x_{i(t)}$ was directly implied to 
$x_{i(t)}^*$ by $\ux^*_{U_{t-1}}$. The set of frozen variables remains the
same, $W_t=W_{t-1}$, as this step is merely the actuation of a previous
logical implication; $(iii)$ $i(t)$ was not directly implied by 
$\ux^*_{U_{t-1}}$. This is the only interesting case that we develop now.

Let us call $Z_t \equiv W_{t}\setminus W_{t-1}$ the set of newly frozen
variables after this fixing step.
A moment of reflection shows that $Z_t$ contains $i(t)$ 
and that it forms
a connected
subset of $V$ in $G$. 
Consider now the subgraph $G_t\subseteq G$ induced by $Z_t$ 
(i.e.  $G_t=(Z_t,F_t,E_t)$ where $F_t$ is the set of factor nodes having 
at least one adjacent variable in $Z_t$, and $E_t$ is the set of edges between 
$Z_t$ and $F_t$). A crucial observation is the following:
\begin{lemma}
If $G_t$ is a tree, no contradiction can arise during 
step $t$.
\end{lemma}
From this lemma, and since the factor graph of a typical random formula 
is locally tree-like, one is naturally lead to study the size 
of $Z_t$, i.e. of the cascade of newly implied variables induced by fixing the
$t$-th variable. If this size remains bounded as $n\to\infty$, then $G_t$ 
will typically be a tree and, consequently, contradictions 
will arise with vanishingly small probability during one step. 
If on the other hand the size diverges for infinitely large samples, 
then $G_t$ will contain loops and opens the possibility for contradictions
to appear.

In order to compute the typical size of $Z_t$, we notice
that $|Z_t|=|W_t|-|W_{t-1}|$, and consider a $t$ of order $n$,
namely $t=n\theta$. If we let $\phi(\theta) \equiv \E|W_{n\theta}|/n$
denote the fraction of frozen variables when a fraction $\theta$ of variables
have been fixed, then under mild regularity conditions we have
\begin{eqnarray}
\lim_{n\to\infty}\E[|Z_{n\theta}|] = \frac{\de\phi(\theta)}{\de\theta}\, .
\end{eqnarray}
Of course $\phi$ will be an increasing function of $\theta$.
The argument above implies that, as long as its derivative remain finite
for $\theta\in [0,1]$, then the algorithm finds a solution. 
When the derivative diverges at some point $\theta_*$, then 
the number of direct implications of a single variable diverges as 
well. The spinodal point 
$\alpha_{\sp}(k)$ is defined as be the smallest value of $\alpha$ such that 
this happens. 

The expectation in the definition of $\phi(\theta)$
is with respect to the choices made by the real BP algorithm in the first 
$n \theta$ steps, including the small mistakes it necessarily makes. 
Our crucial hypothesis is that the location of 
$\alpha_{\sp}(k)$ does not change (in the $n\to\infty$ limit)
if $\phi(\theta)$ is computed along the execution
of the idealized decimation algorithm. In other
words we assume that the cumulative effect of BP errors over $n$
decimation steps produces only a small bias in the distribution
of $\ux^*$. For $\alpha\ge \alpha_{\sBPd}(k)$ this hypothesis is no
longer consistent, as the real BP algorithm fails with high probability.

Under this hypothesis, and recalling the description of the state of the
idealized algorithm  given above, we 
can compute $\phi(\theta)$ as follows. Draw a random formula on $n$
variables, a uniformly random `reference' solution $\ux^*$, a subset $U$ of
$n\theta$ variable nodes\footnote{in the large $n$ limit one can equivalently
draw $U$ by including in it each variable of $V$ independently with
probability $\theta$.}.
Let $\phi_n(\theta)$ be the probability
that a uniformly random variable node $i$ is frozen, 
i.e. either in $U$ or directly implied by $\ux^*_U$. 
Then $\phi(\theta) = \lim_{n\to\infty}\phi_n(\theta)$.
In the next Section this computation will be performed in the random tree 
model $\Tree(\ell)$ of Sec.~\ref{sec:Back}.
%
%
\section{The tree model and its analysis}
\label{sec:Tree}

Let us consider a $k$-satisfiability formula whose factor graph is a finite tree, and the uniform measure $\mu$ over its solutions (which always
exist) defined in Eq.~(\ref{eq:Uniform}).
It follows from general results~\cite{Factor} that the 
recursion equations~(\ref{eq:BPUpdate1},\ref{eq:BPUpdate2}) have a unique 
fixed-point, that we shall denote $\{h_{i \to a}, u_{a \to i} \}$.
Further the BP marginals $\nu_i(\,\cdot\,)$, cf.  Eq.~(\ref{eq:BPMarginal}),
 are the actual marginals of $\mu$.
Drawing a configuration $\ux$ from the law $\mu$ is most easily done in a 
recursive, broadcasting fashion.
Start from an arbitrary variable node $i$ and draw $x_i$ with
distribution $\nu_i$. 
Thanks to the Markov property of $\mu$, conditional on the 
value of $x_i$, $\ux_{V \setminus i}$ can be generated independently for
each of the branches of the tree rooted at $i$. Namely, for each 
$a \in \di$, one draws $\ux_{\da \setminus i}$ from
\begin{equation}
\mu(\ux_{\da \setminus i} | x_i ) = \frac{1}{z} w_a(x_i,\ux_{\da \setminus i})
\prod_{j \in \da \setminus i} \nu_{j \to a}(x_j) \ .
\label{eq_broadcast}
\end{equation}
Here $z$ is a normalization factor and $\nu_{i \to a}(\cdot)$ denotes the
marginal of the variable $x_i$ in the amputated factor graph where factor
node $a$ has been removed (this is easily expressed in terms of the message 
$h_{i \to a}$). 
Once all variables $j$ at distance $1$ from $i$ have been generated, 
the process can be iterated to fix variables at distance $2$ from $i$,
and so on. 
It is easy to realize that this process indeed samples a solution
uniformly at random.

Following the program sketched in the previous Section, 
we shall study the effect of fixing a subset of the variables to
the value they take in one of the solutions. We first state the following
lemma.
\begin{lemma}
Suppose $U$ is a subset of the variables of a tree formula, and let 
$\ux^*$ be a uniformly random solution.
The probability that a variable $i \notin U$ is directly implied by $\ux_U^*$
reads
\begin{equation}
\nu_i(\0t)
\Big\{1 - \underset{a \in \dpi}{\prod}(1 - \ud_{a \to i}) \Big\}
+ \nu_i(\1t)
\Big\{1 - \underset{a \in \dmi}{\prod}(1 - \ud_{a \to i}) \Big\} \ ,
\label{eq:proba_frozen_single}
\end{equation}
where the new messages $\{ \ud_{a \to i},\hd_{i \to a} \}$ are solutions of
\begin{eqnarray}
\hd_{j \to a} &=& \begin{cases} 1 & {\rm if} \ j \in U \\
1 - \underset{b \in \dmj(a)}{\prod}(1 - \ud_{b \to j}) & {\rm otherwise}
\end{cases} 
\label{eq:ud_to_hd_single} \\
\ud_{a\to l} &=& \prod_{j \in \da \setminus l} \left(
\frac{1 - \tanh h_{j \to a}}{2} \, \hd_{j \to a} \right) \ .
\label{eq:hd_to_ud_single}
\end{eqnarray}
\end{lemma}

We consider now a random tree factor graph and 
a random set of fixed variables $U$.

\begin{lemma}
Consider a random tree formula $\Tree(\ell)$ obtained from the construction 
of Section \ref{sec:Back}, and a random subset $U$ of its variable nodes 
defined by letting $j\in U$ independently with probability $\theta$
for each $j$.
Finally, let $\ux^*$ be a uniformly random solution of $\Tree(\ell)$. 
Then the probability that the root of 
$\Tree(\ell)$ is frozen (either fixed or directly implied by $\ux_U^*$) is
\begin{equation}
\phi_\ell^\stree(\theta) = \E_\ell \left[(1-\tanh h) \hd\right] \ ,
\label{eq:phi_ell_tree}
\end{equation}
where $E_\ell[\cdot]$ denotes  expectation with respect to the
distribution of $(h,\hd)_\ell$.
This is a (vector) random variable defined by recurrence on
$\ell$ as
\begin{eqnarray}
\hspace{-5mm} (h,\hd)_\ell \hspace{-3mm} &\ed& \hspace{-3mm}
\left(\sum_{i=1}^{l_+} u_i^+ - \sum_{i=1}^{l_-} u_i^- , \,
1 - \zeta \prod_{i=1}^{l_-} (1-\ud_i^-)\right) \, ,
\label{eq:uud_to_hhd} \\
\hspace{-5mm} (u,\ud)_{\ell+1} \hspace{-3mm} &\ed& \hspace{-3mm}
\left(f(h_1,\dots,h_{k-1}), \,
\prod_{i=1}^{k-1} \frac{1-\tanh h_i}{2}\hd_i\right) \, ,
\label{eq:hhd_to_uud}
\end{eqnarray}
with initial condition $(u,\ud)_{l=0} = (0,0)$ with probability 1.
In this recursion 
$l_+$ and $l_-$ are two independent Poisson random variables of
parameter $\alpha k/2$, $\zeta$ is a random variable equal to $0$ (resp. $1$)
with probability $\theta$ (resp. $1-\theta$), the 
$\{(u_i^+,\ud_i^+),(u_i^-,\ud_i^-)\}$ and $(h_i,\hd_i)$ are independent copies
of, respectively, $(u,\ud)_\ell$ and $(h,\hd)_\ell$.
\end{lemma}

\begin{figure}[t]
\center{\includegraphics[width=7cm]{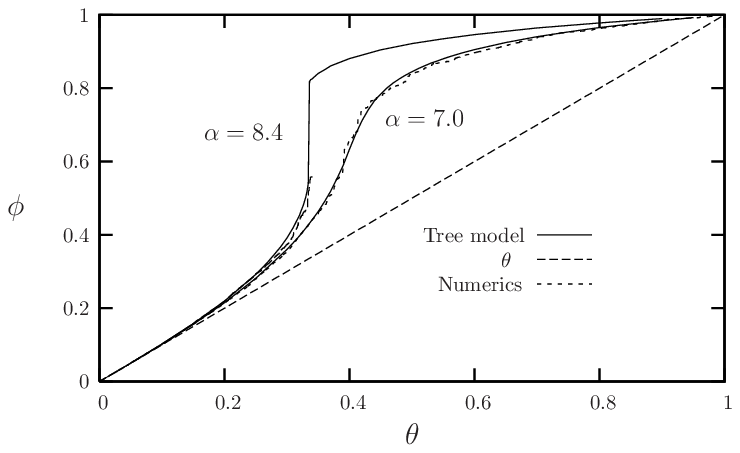}}
\caption{Fraction of frozen variables as a function of the fraction
of fixed variables. Comparison between the tree model and the algorithmic
numerical results, for 4-satisfiability formulas with $n=4000$, $\alpha=7$ and
$\alpha=8.4$.}
\label{fig:phi_of_eta}
\vspace{-2mm}
\end{figure}

To obtain a numerical estimate of the function 
$\phi^\stree(\theta) = \lim_{\ell \to \infty}  \phi_\ell^\stree(\theta)$
we resorted to sampled density evolution (also called `population dynamics' 
in the statistical physics context \cite{Mezard}), using samples of
$10^5$ elements and $k=4$ as a working example, see Fig.~\ref{fig:phi_of_eta}. 
For small values of
$\alpha$, $\phi^\stree(\theta)$ is smoothly increasing and slightly larger
than $\theta$. Essentially all frozen variables are fixed ones, and very few 
directly implied variables appear. Moreover the maximal slope of the curve is 
close to 1, implying that the number of new frozen variables
at each step, $Z_t$, remains close to 1. 
As $\alpha$ grows, $\phi^\stree(\theta)$ becomes 
significantly different from $\theta$, and the maximal slope encountered in
the interval $\theta \in [0,1]$ gets larger. At a value 
$\alpha_{\sp}^\stree(k)$ 
the curve $\phi^\stree(\theta)$ acquires a vertical tangent at 
$\theta_*(\alpha_{\sp}^\stree )$, signaling the divergence of the size of the 
graph of newly implied variables. Density evolution gives us
$\alpha_{\sp}^\stree(k=4)\approx 8.05$, with an associated value of 
$\theta_* \approx 0.35$. For $\alpha>\alpha_{\sp}^\stree(k)$ the
curve $\phi^\stree(\theta)$ has more than one branch,
corresponding to the presence of multiple fixed points
for $\theta\in [\theta_0(\alpha),\theta_*(\alpha)]$.
In analogy with \cite{Maxwell2}, we expect the evolution of the algorithm 
to be described by picking (for each $\theta$) 
the lowest branch of $\phi^\stree(\theta)$. The resulting curve has 
a discontinuity at $\theta_*(\alpha)$, which is a  slowly decreasing function
of $\alpha$.

We expect the tree 
computation to provide the correct prediction for the actual curve
$\phi(\theta)$ (i.e. $\phi^\stree(\theta) = \phi(\theta)$) 
for a large range of the satisfiable regime, including 
$[0,\alpha_{\sp}^\stree(k)]$. As a consequence, we expect 
$\alpha_{\sp}(k)=\alpha_{\sp}^{\stree}(k)$ and BP decimation to be successful
up to $\alpha_{\sp}(k)$.
Similar tree computations are at the basis of a number of
statistical mechanics computations in random $k$-SAT and have been repeatedly
confirmed by rigorous studies.

The relation between tree and graph can be formalized 
in terms of  Aldous~\cite{Aldous} local weak convergence method.
Fix a finite integer $\ell$ and
consider the finite neighborhood $\Ball(\ell)$ of radius $\ell$ around an
arbitrarily chosen variable node of an uniformly drawn factor graph $G$ on
$n$ variables.  Denote by
$\mu_{\Ball(\ell),n}(\, \cdot \,)$ the law of $\ux_{\Ball(\ell)}$ 
when $\ux$ is a uniformly random solution.
We proceed similarly in the random tree ensemble. Draw a random 
tree $\Tree(L)$ with $L \ge \ell$, let $\Tree(\ell)$ its first $\ell$
generations, and  
$\mu_{\Tree(\ell),L}^\stree(\, \cdot \,)$ the distribution of 
$\ux_{\Tree(\ell)}$. 
Considerations building on the field of statistical mechanics of
disordered systems leads to the following hypothesis.
\begin{conj}
There exists a sequence $\alpha_{\rm c}(k)$ such that
$\mu_{\Tree(\ell)}(\, \cdot \,) \ed \mu_{\Tree(\ell)}^\stree(\, \cdot \,)$
for all $\alpha < \alpha_{\rm c}(k)$, i.e. 
$(\Ball(\ell),\mu_{\Ball(\ell),n}(\, \cdot \,))$ and
$(\Tree(\ell),\mu_{\Tree(\ell),L}^\stree(\, \cdot \,))$ have the same
weak limit. A precise determination of 
$\alpha_{\rm c}(k)$ was presented in~\cite{OurPNAS}, yielding 
$\alpha_{\rm c}(k) \approx 3.86 \, , 9.55 \, , 20.80 $ for, respectively,
$k=3 \, , 4 \, , 5$, and $\alpha_{\rm c}(k) = 2^k \log 2 - \frac{3}{2} \log 2 +
O(2^{-k})$ at large $k$.
\end{conj}
Local weak limits of combinatorial models on random graphs were
recently considered in~\cite{Antoine}. For a generalized conjecture
in the regime $[\alpha_{\rm c}(k),\alpha_{\rm s}(k)]$ see \cite{rearr}.

A slightly stronger version of this conjecture would imply that
$\phi(\theta)= \phi^{\stree}(\theta)$. As a consequence (following
the discussion in previous section) the tree model would correctly describe
the algorithm evolution.

%
%
\section{Numerical simulations}
\label{sec:Numerical}

In order to test the validity of our analysis we performed
 numerical simulations of the pseudo-code of Table \ref{tab:Algo}.
Let us give a few further details on its implementation. The BP messages
are stored as $\{\tanh h_{i \to a},\tanh u_{a \to i} \}$. Ambiguities in
the update rule (\ref{eq:BPUpdate2}) arises when 
$\tanh u_{b \to i} = \tanh u_{c \to i} =1$ with $b \in \dpi(a)$ and 
$c \in \dmi(a)$. Because of numerical imprecisions this situation can occur 
even before a contradiction has been detected by WP; such ambiguities are
resolved by recomputing the incoming messages $\tanh u_{b \to i}$ using
the regularized version of Eq.~(\ref{eq:FDefinition}) with a small positive
value of $\epsilon$ (in practice we used $\epsilon=10^{-4}$).

As for the stopping criterion used in step 5, we leave
the BP iteration loop if either of the two following criteria is fulfilled:
$(1)$, $\sup_i |\tanh h_i^{(r)} - \tanh h_i^{(r-1)}|< \delta$, i.e. BP has 
converged to a fixed-point within a given accuracy; $(2)$ A maximal number
of iterations $r_{\rm max}$ fixed beforehand has been reached. In our
implementation we took $\delta = 10^{-10}$ and $r_{\rm max}=200$.

A first numerical check is presented in Fig.~\ref{fig:phi_of_eta}. The two
dashed curves represent the fraction of frozen variables along the
execution of the BP guide decimation algorithm, for two formulas of the
$4$-sat ensemble, of moderate size ($n=4000$). The first formula
had a ratio of constraints per variable $\alpha=7<\alpha_{\sp}$. In agreement
with the picture obtained from the analytical computation, the algorithm
managed to find a solution of the formula (no contradiction encountered) and
the measured fraction of frozen variables follows quite accurately the tree
model prediction. The second formula was taken in the regime 
$\alpha_\sp < \alpha =8.4 < \alpha_{\rm c}$. The algorithm halted
because a contradiction was found, after roughly the fraction
$\theta_*$ (computed from the tree model) of variables has been fixed. The
portion of the curve before this event exhibits again a rather good agreement
between the direct simulation and the model.

\begin{figure}[t]
\center{\includegraphics[width=7cm]{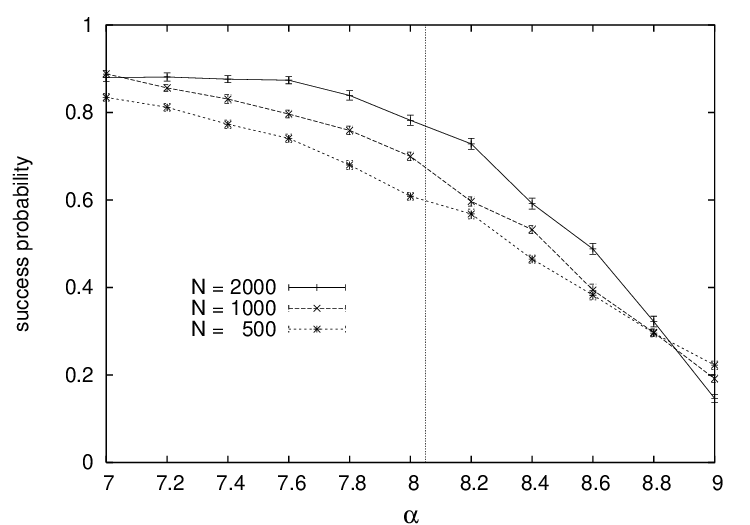}}
\caption{Probability of success of the BP decimation algorithm 
as a function of the clause density $\alpha$ in random $4$-SAT.
The vertical line corresponds to the threshold $\alpha_{\sp}(4)$.
Our analysis indicates that BP decimation finds a solution
with probability bounded away from $0$ for $\alpha<\alpha_{\sp}(4)$.}
\label{fig:BPconvProb}
\vspace{-2mm}
\end{figure}

Figure \ref{fig:BPconvProb} shows the probability of success of
BP decimation in a neighborhood of $\alpha_{\sp}(4)$
for random formulae of size $n=500$, $1000$, $2000$. 
Each data point is obtained by running the algorithm on 
$1000$ to $3000$ formulae. The data strongly suggest that the 
success probability is bounded away from $0$ for 
$\alpha<\alpha_\sp(k)$, in agreement with our argument.
\begin{figure}[t]
\center{\includegraphics[width=7cm]{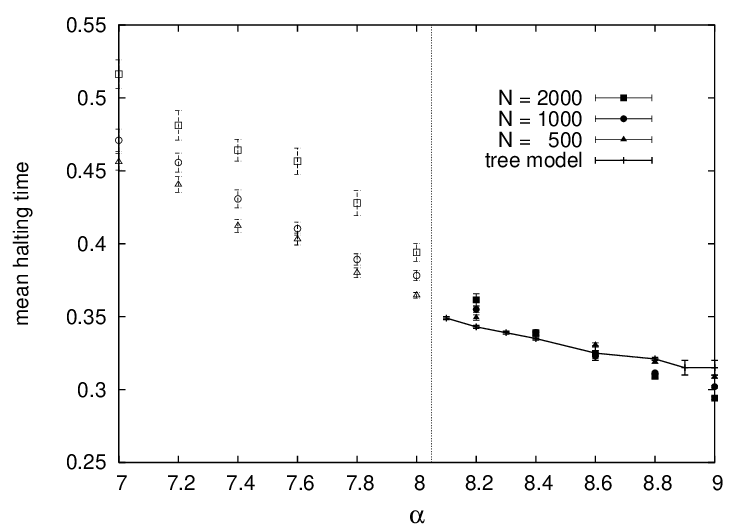}}
\caption{Mean halting time for the BP decimation algorithm
 in random $4$-SAT.  
The vertical  line corresponds to the threshold $\alpha_{\sp}(4)$.
The mean is taken over unsuccessful runs.
For  $\alpha<\alpha_{\sp}(4)$ a large fraction of the runs 
is successful and do not contribute to the mean.}
\label{fig:haltTime}
\vspace{-2mm}
\end{figure}

Finally, in Figure \ref{fig:haltTime} we consider the number of variables 
$t_*$ fixed by BP decimation before a contradiction is encountered.
According to the argument in Section \ref{sec:Intuitive}, $t_*/n$ should
concentrate around the location $\theta_*$ of the 
discontinuity in  $\phi(\theta)$. This is in fact the point
at which the number of variables directly implied by a fixed one
is no longer bounded. The comparison is again encouraging.
Notice that for $\alpha<\alpha_\sp(k)$ we do not have any
prediction, and the estimate of $t_*$ concerns only a small fraction of 
the runs.

To summarize, our simulations support the claim that, for
$\alpha<\alpha_\sp(k)$ 
the success probability is strictly positive and the algorithm evolution follows the tree model. For $\alpha>\alpha_{\sp}(k)$ the main failure mechanism
is indeed related to unbounded 
cascades of directly implied variables, 
after about $n\theta_*$ steps.

%
%
\section{Conclusions and future directions}
\label{sec:Conclusions}

Let us conclude by highlighting some features of this work and proposing some
directions for future research. It is worth mentioning that, as was also found
in~\cite{OurPNAS}, random 3-sat has a qualitatively different behavior
compared to random $k$-sat with $k \ge 4$. In particular we did not found any
evidence for the existence of a vertical tangent point in the $k=3$ function
$\phi(\theta)$ in the regime we expect to control through the tree
computation, namely $\alpha < \alpha_{\rm c}(3) \approx 3.86$.

Our analysis suggests that BP guided decimation is successful
with positive probability  for $\alpha\le \alpha_{\sp}(k)$. 
Further we argued that this threshold can be computed through
a tree model and evaluated via density evolution.   
Despite these conclusions are  based on several assumptions, it is 
tempting to make a comparison with the best rigorous 
results on simple decimation algorithms.
For $k=4$ the best result was obtained by Frieze and Suen~\cite{Frieze}
who proved SCB (shortest clause with limited amount of backtracking) 
to succeed for $\alpha < 5.54$. This is far from 
the conjectured threshold of BP decimation that is  $\alpha_{\sp}(4)\approx
8.05$.
For large $k$, an asymptotic expansion suggests that
\begin{equation}
\alpha_{\sp}(k) = e \frac{2^k}{k} (1+O(k^{-1})) \ ,
\label{eq:conj_thr}
\end{equation}
whereas SCB is known from~\cite{Frieze} to reach clause densities of 
$c_k 2^k /k$, with $c_k \to 1.817$ as $k \to \infty$. A rigorous version of
our analysis would lead to a constant factor improvement.
On the other hand, the quest for an algorithm that provably solves
random $k$-SAT in polynomial time beyond $\alpha = O(2^k / k)$, is open.

From a practical point of view the decimation strategy studied in this paper
is not the most efficient one. A seemingly slight modification of the
pseudo-code of Table~\ref{tab:Algo} consists in replacing the uniformly
random choice of the variable to be fixed, privilegiating the ones with the
most strongly biased marginals. The intuition for this choice is that these
marginals are the less subject to the 'small errors' of BP. The numerical
results reported in~\cite{OurPNAS} suggest that this modification improves
significantly the performances of the decimation algorithm; unfortunately it
also makes its analysis much more difficult.

This work was partially supported by EVERGROW, integrated project No. 1935 
in the
complex systems initiative of the Future and Emerging Technologies directorate 
of the IST Priority, EU Sixth Framework.

%
%


\appendix

\begin{proof}[Proof of Lemma 1]
The statement is completely analogous to the equivalence between message 
passing and peeling versions of erasure decoding for LDPC codes
\cite{MCT}. 
Since the proof follows the same lines as well, we will limit
ourselves to sketch its main points. 

1) Let  $\{\uw_{a\to i}\}$ and $\{\uw_{a\to i}'\}$
be two fixed points of WP. Then  
$\{\min(\uw_{a\to i},\uw_{a\to i}')\}$ is a fixed point
as well. It follows that the `minimal' fixed point is well defined and
that it coincides with the limit of $\{\uw_{a\to i}^{(r)}\}$ 
irrespective of the order of WP updates.

2) Consider the ordering $\{i(1),i(2),\dots,i(q)\}$ according to 
which variables are declared as directly implied within UCP.
For each $s\in \{1,\dots,q\}$ there is at least one unit clause 
involving only variable $i(s)$ before this was declared. Call this 
$a(s)$. Then use the same update order for WP, namely
update, in sequence message $\uw_{a(s)\to i(s)}$, and all the messages
$\hw_{i(s)\to b}$ for $b\neq a(s)$. It is immediate to show
that this leads to a fixed point, and the  resulting
WP-implied variables coincide with the directly implied variables.
The proof is completed by using point $1$.

3) Consider the same ordering of variables used in point 2 above.
If there exists $i\in V$, $a\in \dpi$, $b\in\dmi$ as in the statement, 
then UCP must have reduced both clauses $a$ and $b$ to a unit clause 
involving $x_i$ and requiring it to take different values. Viceversa
if UCP produces such a situation, in the WP updates 
$\uw_{a\to i}^{(r)}=\uw_{b\to i}^{(r)}=\imp$ after some time $r$.
\end{proof}

\begin{proof}[Proof of Lemma 2]
The same statement has been proved for the Maxwell decoder
\cite{Maxwell2}. We therefore briefly recall the basic ideas used in 
that case. 

First of all the only WP messages  changing from step $t-1$
to step $t$  (call these the `new' messages) are the ones on the edges 
of the tree $G_t$, and directed outwards. 
As a consequence, no contradiction can
arise because of two contradicting new messages, because no
variable node has two incoming new messages. 

There could  be, in line of principle, a contradiction between a new and 
an old message.
The crucial observation is that indeed any factor node
in $F_t$  has at most two adjacent variable nodes in $Z_t$
(because otherwise if could not `transmit' an implication).
If a variable node $i$ already receives some $\imp$
message at time $t-1$ from clause $a$, then  it cannot receive any 
new message at time $t$ from a different clause $b$. This because 
the message $i\to b$ must already be $\imp$, and therefore 
clause $b$ is already effectively `reduced'. 

An alternative argument consists in considering the equivalent UCP
representation. If $G_t$  is a tree, then no variable appears twice 
in a unit clause, and therefore no contradiction arises.
\end{proof}

\begin{proof}[Proof of Lemma 3]
Since we are dealing with a tree graph, equations 
(\ref{eq:ud_to_hd_single},\ref{eq:hd_to_ud_single}) admit
a unique solution, determined from the boundary condition $\hd_{i \to a}=1$
(resp. $\hd_{i \to a}=0$) if $i$ is a leaf in $U$ (resp. a leaf outside of
$U$). The newly introduced messages have the following interpretation. 
Imagine running WP, cf. Eqs.~ (\ref{eq:WPUpdate1}), 
(\ref{eq:WPUpdate2}) to find which variables are directly implied by $\ux^*_U$.
Then $\ud_{a \to j}$ is the probability that $\uw_{a \to j}=\imp$
when $\ux^*_U$ is drawn conditional on $x_j$ satisfying $a$.
Further, $\hd_{j \to a}$ is the 
probability that $\hw_{j \to a}=\imp$ when $\ux^*_U$ is drawn conditional on 
$x_j$ not satisfying clause $a$.

Now, suppose $x_i$ has
been fixed to $x_i^*$ drawn according to its marginal (hence the two terms in
Eq.~(\ref{eq:proba_frozen_single})) and a configuration $\ux$ has been
generated conditional on $x_i$, through the broadcast construction. 
Then the configuration of
the variables in $U$ is retained, $\ux_U=\ux_U^*$, and the rest of $\ux^*$
is discarded. The
status (directly implied or not) of $x_i$ is read off from the values of
the messages $\uw_{a \to i}$ it receives. It is easy to convince oneself
that $x_i$ cannot be implied to take the value opposite to the one it
took at the beginning of the broadcasting: by definition $\ux_U^*$ is
compatible with it. Equation (\ref{eq:proba_frozen_single}) follows 
by computing the probability that at least one of the messages 
$\uw_{a\to i}$ is equal to $\imp$ among the ones from clauses $a$
that are satisfied by $x_i^*$.

Equation (\ref{eq:ud_to_hd_single}) is derived by applying the same argument 
to the branch of the tree rooted at $j$ and not including
factor node $a$. Finally, to derive 
Eq.~(\ref{eq:hd_to_ud_single}) notice that, in order for variable
$x_l$ to be directly implied to satisfy clause $a$, each of the variables
$j\in\da\setminus l$ must be implied by the corresponding 
subtree not to satisfy $a$. From the above remark, this
can happen only if  none of the $\{x^*_j\}$ satisfies $a$.
The probability of this event is easily found from (\ref{eq_broadcast}) to be 
\begin{equation}
\prod_{j \in \da \setminus i} \frac{1 - \tanh h_{j \to a}}{2} \ .
\end{equation}
\end{proof}

\begin{proof}[Proof of Lemma 4]
Denote by $\rho$ the root of $\Tree(\ell)$.
Conditional on the realization of  the tree and of the set $U$, the probability
of a direct implication of the root is obtained by solving
(\ref{eq:BPUpdate1}), (\ref{eq:BPUpdate2}),
(\ref{eq:ud_to_hd_single}), (\ref{eq:hd_to_ud_single})
for the edges directed towards the root, which leads to couples of messages 
$\{(h_{i \to a},\hd_{i \to a})\}$ and $\{(u_{a \to i},\ud_{a \to i})\}$
along the edges of $\Tree(\ell)$.
Since $\Tree(\ell)$ and $U$ are random 
these couples of messages are random variables as well. 

We claim that for 
$\ell \ge 1$, the messages $(u_{a \to \rho},\ud_{a \to \rho})$ 
sent to the root of
$\Tree(\ell)$ by the adjacent constraint nodes are distributed as 
$(u,\ud)_\ell$. Similarly for $\ell \ge 0$, 
$(h,\hd)_\ell$ has the distribution of the messages 
sent from the first generation variables to their ancestor constraint node
in a random $\Tree(\ell+1)$. This claim is a direct consequence of
Eqs.~(\ref{eq:BPUpdate1}), (\ref{eq:BPUpdate2}), (\ref{eq:ud_to_hd_single}),
(\ref{eq:hd_to_ud_single})
and of the definition of $\Tree(\ell)$ and $U$. The random variables $l_\pm$
have, for instance, the distribution of the cardinalities of $\dpmi(a)$ for an
arbitrary edge of the random tree, as 
$|\di \setminus a|\ed\poisson(\alpha k)$ and unsatisfying values $z(i,a)$
of the variables are chosen independently with equal probability. 

Finally
the expression of $\phi_\ell^\stree(\theta)$ is obtained from
(\ref{eq:proba_frozen_single}) by noting that the cardinalities of 
$\dpmi$ for the root of $\Tree(\ell)$ are distributed as the ones of 
$\dpmi(a)$ and using the global symmetry between $\0t$ and $\1t$, which
implies that on average the two terms of (\ref{eq:proba_frozen_single}) yield
the same contribution. Note that the dependence on $\theta$ of 
$\phi_\ell^\stree$ arises through the
distribution of $(h,\hd)_\ell$, the bias of the coin $\zeta$ used
in (\ref{eq:uud_to_hhd}) being $\theta$.
\end{proof}

\begin{proof}[Details on the population dynamics algorithm]
The numerical procedure we followed in order to determine 
$\phi_\ell^\stree(\theta)$ amounts to  
approximating the distribution of the random variable $(u,\ud)_\ell$ 
by the empirical distribution of a large sample of couples 
$\{(u_j,\ud_j)\}_{i=1}^N$. 
A sample $\{(h_j,\hd_j)\}_{i=1}^N$
is then generated according to Eq.~(\ref{eq:uud_to_hhd}): for
each $j \in [N]$ one draws two Poisson random variables $l_+$ and $l_-$,
$l_+ + l_-$ indexes $j_i^\pm$ uniformly in $[N]$, and a biased coin $\zeta$.
The $j$-th element of the sample is thus computed as
\begin{equation}
(h_j,\hd_j)= \left(\sum_{i=1}^{l_+} u_{j_i^+} - \sum_{i=1}^{l_-} u_{j_i^-} , \,
1 - \zeta \prod_{i=1}^{l_-} (1-\ud_{j_i^-})\right) \ . \nonumber
\end{equation}
Subsequently the sample $\{(u_j,\ud_j)\}$ is updated from $\{(h_j,\hd_j)\}$
by a similar interpretation of Eq.~(\ref{eq:hhd_to_uud}). 
After $\ell$ iterations
of these two steps, starting from the initial configuration
$(u_j,\ud_j)=(0,0)$ for all $j \in [1,N]$, the estimate of 
$\phi_\ell^\stree(\theta)$
is given by
\begin{equation}
\frac{1}{N} \sum_{j=1}^N (1-\tanh h_j) \hd_j \ .
\end{equation}
When $\ell$ gets large this quantity is numerically found to converges to a 
limit we denoted $\phi^\stree(\theta)$. 
\end{proof}

\newpage

\begin{proof}[Large $k$ argument]
Consider the function $\hphi(\theta)$ defined, for $\theta \in [0,1]$,
as the smallest solution in $[0,1]$ of the equation
\begin{equation}
\hphi = \theta + (1-\theta)
\left(1-\exp\left[-\frac{\alpha k}{2^k} \hphi^{k-1}\right] \right) \ . 
\label{eq:hphi}
\end{equation}
It can be shown that $\hphi(\theta)$ is a smoothly increasing
function of $\theta$ as long as $\alpha < \halpha_{\sp}(k)$, while for
larger values of $\alpha$ a discontinuous jump develops in its curve. This
threshold can be explicitly computed and reads
\begin{equation}
\halpha_{\sp}(k) = 
\frac{2^k}{k} \left(\frac{k-1}{k-2}\right)^{k-2} \ .
\label{eq:halpha}
\end{equation}
We believe this simple to determine function $\hphi(\theta)$ to be equivalent
to the true $\phi(\theta)$ in the large $k$ limit, up to exponentially small
in $k$ corrections. In fact  
(\ref{eq:phi_ell_tree},\ref{eq:uud_to_hhd},\ref{eq:hhd_to_uud})
implies the following exact equation,
\begin{equation}
\E[\hd] = \theta + (1-\theta) 
\left(1 - \exp\left[-\frac{\alpha k}{2^k} \phi(\theta)^{k-1} \right] \right) 
\ ,
\label{eq:hd_exact}
\end{equation}
where the expectation is taken in the $\ell \to \infty$ limit. For large
values of $k$ one can show the random variable $h$ to be exponentially close 
to $0$, hence $\phi(\theta)$ and $\E[\hd]$ coincide at the leading order, and
by comparing (\ref{eq:hphi}) and (\ref{eq:hd_exact}) they also coincide
with $\hphi(\theta)$. The conjecture stated in Eq.~(\ref{eq:conj_thr})
was obtained by expanding (\ref{eq:halpha}) at the leading order.
\end{proof}

\end{document}